\documentclass[11pt]{article}
\usepackage[margin=1in]{geometry}

\usepackage{amsmath,amssymb}
\usepackage{booktabs}
\usepackage{enumitem}
\usepackage{graphicx}
\usepackage[hypertexnames=false]{hyperref}
\usepackage{microtype}
\usepackage{tikz-cd}
\usepackage{multicol}
\usepackage{float}
\usepackage{tabularx}
\usepackage{array}

\title{Active Learning Guided Design Space Refinement for Scalable Multi-Objective Bayesian Optimization in Materials Discovery}

\author{%
Ntagiantas A.\textsuperscript{+}, Tsilimidos P.\textsuperscript{+}, Giannakopoulos G., Rekatsinas C.\textsuperscript{*}, Krokidas P.\textsuperscript{*} \textsuperscript{\S}\\
\small National Centre for Scientific Research ``Demokritos''\\
\small \textsuperscript{+}Shared first authorship; \textsuperscript{*}Shared last authorship; \textsuperscript{\S}Corresponding Author: \href{mailto:p.krokidas@iit.demokritos.gr}{p.krokidas@iit.demokritos.gr}
}
\date{\today}

\begin{document}
\maketitle

\begin{abstract}
Advanced materials discovery increasingly relies on machine learning and Bayesian optimization to explore large discrete design spaces under limited evaluation budgets. However, conventional Bayesian optimization (BO) can become inefficient as candidate spaces grow, often evaluating low-value regions before reaching informative areas. We propose an active-learning (AL)-guided adaptive search-space refinement framework combined with multi-objective BO to accelerate materials optimization while preserving Pareto-relevant regions. We evaluate the approach on CH$_4$/N$_2$ separation in covalent--organic frameworks and pressure-vessel design with material-direction stress components and thickness objectives. Results show that the AL-guided refinement reduces the candidate space by approximately half while preserving more than 99 percent of the original hypervolume. The reduced-space strategy improves early convergence and cumulative Pareto-front discovery from the BO, demonstrating efficient large-scale materials optimization across constrained autonomous materials discovery settings.
\end{abstract}

\section{Introduction}

The rapid expansion of computational materials databases and high-throughput simulation techniques has fundamentally transformed the discovery and optimization of advanced materials. Machine learning (ML) and data-driven methodologies now enable the exploration of increasingly large design spaces across applications including gas separation, catalysis, energy storage, and adsorption-based processes. At the same time, the growing complexity and dimensionality of modern materials datasets introduce significant challenges for efficient optimization, particularly when candidate evaluation remains computationally demanding. As accessible materials spaces continue to expand, identifying promising candidates under limited computational budgets has become a major challenge in autonomous materials discovery.

Bayesian optimization (BO) has emerged as an effective approach for data-efficient optimization in materials science\cite{ref1,ref6,ref14} because it can guide exploration using probabilistic surrogate models and uncertainty-aware acquisition strategies. Unlike conventional optimization techniques, BO balances exploration and exploitation while operating under restricted evaluation budgets, making it attractive for expensive scientific optimization problems. Recent applications have demonstrated the potential of BO to accelerate nanoporous-material screening, molecular discovery, adsorption optimization, and multi-objective materials engineering\cite{ref12,ref17,ref21,ref23,Loutas2025,krokidas2026frugalbayesianoptimizationscalable} while reducing the number of computationally expensive evaluations required to identify promising solutions\cite{ref18,ref24,ref34}.

Despite these advantages, the scalability of conventional BO remains a major limitation in ultra-large design spaces. As the number of candidate materials increases, the optimization process often becomes progressively less efficient and may require many evaluations before converging toward high-performing regions. In practice, a significant fraction of computational effort can be allocated to low-value candidates that contribute minimally to the final optimization outcome. This issue becomes even more pronounced in multi-objective settings, where competing objectives must be balanced while maintaining diversity across the Pareto front\cite{ref2,ref8,ref9,ref10,ref30,ref31}. Consequently, the computational cost associated with large-scale optimization can become prohibitive, limiting the practical applicability of conventional BO frameworks in autonomous materials discovery pipelines. To improve optimization efficiency in large-scale scientific problems, recent research has increasingly explored the integration of active learning strategies with adaptive exploration mechanisms. By iteratively identifying informative candidate regions, active learning approaches can reduce redundant evaluations and focus computational resources on areas that are more likely to contain high-performing solutions\cite{ref4,ref5,ref7,ref24,ref25,ref26,ref27}. In materials optimization, adaptive refinement strategies are particularly appealing because they can reduce the effective search space before expensive optimization stages. Such approaches may substantially reduce unnecessary exploration in large candidate spaces.

Nevertheless, existing workflows often emphasize either predictive modeling or optimization performance independently, without explicitly incorporating adaptive search-space refinement as a central component of the optimization process. Many large-scale optimization frameworks continue to rely on full-space exploration strategies that become increasingly inefficient as candidate spaces and objective complexity grow. This limitation is particularly relevant in modern materials discovery settings, where optimization frequently involves competing objectives, expensive evaluations, and large candidate pools derived from high-throughput computational screening pipelines.

Inspired by recent works that deal with narrowing the initial search space of BO methods \cite{zombi, focalbo}, in this work, we introduce an adaptive search-space refinement framework for improving the scalability and search focus of Bayesian optimization in materials discovery applications\cite{ref11,ref12,ref17,ref18,ref20,ref21,ref22,ref23}. The framework uses active learning as the candidate-space partitioning component, specifically a density-aware variant developed by some of the authors, density-aware greedy sampling (DAGS)\cite{dags}; the subsequent reduction and optimization strategy is developed here to support multi-objective materials search. In this workflow, active-learning-informed partitions are used to retain candidate regions enriched in high-performing and Pareto-relevant materials, and the resulting reduced space is then passed to Bayesian optimization as a focused warm-start search space. This allows expensive evaluations to be concentrated on promising trade-off regions rather than distributed uniformly across the full design space. We assess the methodology across two optimization problems: CH$_4$/N$_2$ separation in covalent--organic frameworks under PSA operating conditions and pressure-vessel design involving material-direction stress components and thickness objectives. Performance is evaluated in terms of convergence behavior, optimization quality, search-space reduction, and Pareto-front discovery. The results demonstrate that adaptive search-space refinement can reduce the candidate-space size while maintaining competitive Pareto-front quality across diverse materials optimization tasks, highlighting its potential for focused large-scale materials discovery workflows.

\section{Methodology}

\subsection{Problem Formulation}

The optimization problems considered in this work are formulated as expensive black-box materials optimization tasks\cite{ref1}, where the objective functions are not analytically accessible and each evaluation may require significant computational resources. Let $\mathcal{X} \subset \mathbb{R}^d$ denote the design space of candidate materials described by $d$-dimensional feature vectors, and let $f(x)$ represent the objective function associated with a candidate material $x \in \mathcal{X}$. 

\textbf{Single-objective optimization.} In the single-objective setting, the goal is to identify the material configuration that maximizes the target property under a limited evaluation budget,
\begin{equation}
    x^* = \arg\max_{x \in \mathcal{X}} f(x),
\end{equation}
while minimizing the number of costly evaluations required during optimization.

\textbf{Multi-objective optimization.} For multi-objective optimization problems, the objective extends to simultaneously optimizing multiple competing properties,
\begin{equation}
    \mathbf{f}(x) = \left[f_1(x), f_2(x), \ldots, f_m(x)\right],
\end{equation}
where $m$ denotes the number of objectives. Pareto dominance is defined as
\begin{equation}
    x^{(a)} \succ x^{(b)} \Longleftrightarrow
    \left(\forall j \in \{1, \ldots, m\},\ f_j(x^{(a)}) \geq f_j(x^{(b)})\right)
    \land
    \left(\exists j : f_j(x^{(a)}) > f_j(x^{(b)})\right).
\end{equation}
The Pareto-optimal set is therefore
\begin{equation}
    \mathcal{P} = \{x \in \mathcal{X} \mid \nexists x' \in \mathcal{X}: x' \succ x\}.
\end{equation}
In such settings, the optimization process aims to identify Pareto-optimal solutions that achieve favorable trade-offs among competing objectives rather than a single globally optimal point. This formulation is particularly relevant in materials discovery applications, where improving one property often degrades another, resulting in complex objective landscapes and highly nonlinear trade-offs. The proposed framework is designed for large-scale optimization scenarios in which the candidate space may contain tens of thousands of possible material configurations. Under these conditions, exhaustive exploration becomes computationally impractical, motivating adaptive strategies that focus the search process on informative and high-value regions of the design space.

\subsection{Bayesian Optimization}

Bayesian optimization\cite{ref1,ref6} was employed as the primary optimization workflow because of its ability to operate efficiently under limited evaluation budgets while accounting for uncertainty during search. The optimization procedure relies on a probabilistic surrogate model that approximates the underlying objective function using previously evaluated samples. At each iteration, the surrogate model is updated with newly acquired observations and then used to guide the selection of future candidate evaluations.

Gaussian process (GP) models were used as surrogate functions\cite{ref6} because they provide predictive uncertainty estimates alongside objective predictions. Given a set of evaluated samples, the GP surrogate models the posterior distribution of the objective function and enables the computation of acquisition functions that balance exploration and exploitation:
\begin{align}
    \mathcal{D}_t &= \{(x_i, y_i)\}_{i=1}^{t}, \\
    f(x) &\sim \mathcal{GP}\left(\mu(x), k(x, x')\right), \\
    f(x) \mid \mathcal{D}_t &\sim \mathcal{N}\left(\mu_t(x), \sigma_t^2(x)\right).
\end{align}
For single-objective optimization tasks, expected improvement (EI)\cite{ref6} was used to prioritize candidate points with high expected performance relative to the current best observation. The Expected Improvement acquisition function is defined as
\begin{equation}
    \mathrm{EI}(x) = \mathbb{E}\left[\max\left(f(x) - f^{+}, 0\right)\right],
\end{equation}
where $f^{+}$ denotes the best observed objective value.

In the multi-objective setting, acquisition strategies were designed to improve Pareto-front discovery while maintaining diversity among candidate solutions. For multi-objective optimization, the Hypervolume Improvement\cite{ref2,ref9,ref10} is defined as
\begin{equation}
    \mathrm{HVI}(x) = \mathrm{HV}\left(\mathcal{P}_t \cup \{x\}\right) - \mathrm{HV}\left(\mathcal{P}_t\right),
\end{equation}
where $\mathcal{P}_t$ denotes the current Pareto set approximation. In particular, the optimization process focused on maximizing hypervolume improvement to efficiently explore trade-offs among competing objectives.
In all reported multi-objective experiments, candidate selection was implemented using the batch noisy expected hypervolume improvement acquisition function, qNEHVI, as provided by BoTorch. The same acquisition strategy and candidate-pool size were used for both the full-space and reduced-space BO configurations.

The next candidate is selected according to
\begin{equation}
    x_{t+1} = \arg\max_x \alpha(x),
\end{equation}
where $\alpha(x)$ denotes the acquisition function. In practice, the optimization proceeds iteratively by updating the Gaussian process surrogate after each expensive objective evaluation. The posterior predictive distribution is subsequently used to optimize the acquisition function, balancing exploration of uncertain regions and exploitation of high-performing candidates. This sequential decision-making process continues until the predefined evaluation budget is exhausted. Overall, the BO workflow proceeds as observations $\rightarrow$ GP surrogate $\rightarrow$ posterior prediction $\rightarrow$ acquisition evaluation $\rightarrow$ next candidate point.

\begin{figure}[t]
    \centering
    \includegraphics[width=0.95\textwidth]{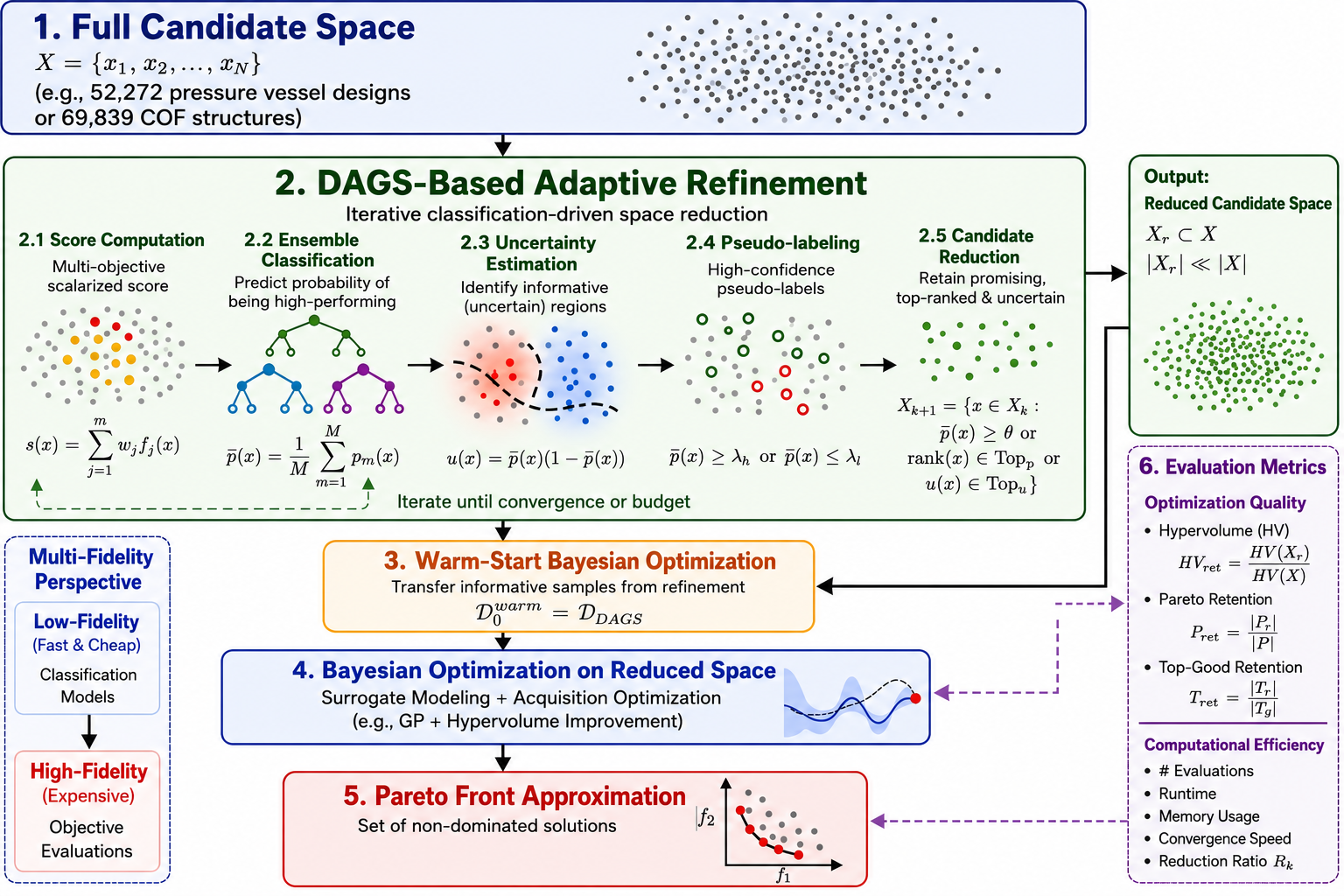}
    \caption{Overall workflow of the proposed DAGS-guided Bayesian optimization framework. The diagram summarizes adaptive refinement, informative sample transfer, and warm-start Bayesian optimization.}
    \label{fig:dags-guided-bo-framework}
\end{figure}

Although BO is highly sample-efficient compared with conventional optimization methods, its performance can deteriorate as candidate-space size increases\cite{ref14,ref21,ref23}. In large-scale materials optimization problems, many evaluations may be allocated to low-performing candidate regions before convergence toward informative areas is achieved. This limitation motivates the integration of adaptive search-space refinement within the optimization workflow. Alternative global optimization algorithms, such as Particle Swarm Optimization (PSO) and Genetic Algorithms (GA), have also been applied in expensive optimization problems. However, unlike Bayesian optimization, these approaches do not explicitly exploit probabilistic surrogate uncertainty, resulting in substantially larger evaluation budgets in high-cost optimization scenarios\cite{ref14}. The overall DAGS-guided refinement and warm-start BO workflow is summarized in Figure~\ref{fig:dags-guided-bo-framework}.

\subsection{DAGS-Based Space Refinement}

To improve optimization scalability, an adaptive DAGS-based refinement strategy was incorporated before the main BO stage. The primary objective of this component is to progressively learn a more accurate partition of the search space while preserving regions that are likely to contain high-performing candidate solutions. At each refinement iteration $k$, the candidate space evolves as
\begin{equation}
    \mathcal{X}_k \subset \mathcal{X}_{k-1},
\end{equation}
with
\begin{equation}
    |\mathcal{X}_k| < |\mathcal{X}_{k-1}|.
\end{equation}
The corresponding reduction ratio is defined as
\begin{equation}
    R_k = 1 - \frac{|\mathcal{X}_k|}{|\mathcal{X}_0|},
\end{equation}
which summarizes the progression from the full space $\mathcal{X}_0$ to the reduced space $\mathcal{X}_k$ through a measurable reduction metric. Rather than optimizing directly over the full candidate space, the proposed methodology first applies an iterative filtering procedure that partitions the design space into informative and non-informative regions using classification-based active learning\cite{ref4,ref5,ref7}. Candidate samples predicted to belong to low-performing regions are discarded only after completion of the refinement stage, when the final classifier is applied to the entire candidate space to construct the reduced search space.

The refinement process operates iteratively using a progressively expanding labeled set. After completion of the refinement iterations, the final classifier is applied once to the entire candidate space in order to construct the reduced search space. At each refinement iteration, a single XGBoost classifier is trained on the progressively expanding labeled set to distinguish promising from non-promising candidate regions. The classifier produces a probability estimate for every candidate in the current pool, which is subsequently used for uncertainty-guided querying, pseudo-label assignment, and reduced-space construction.

This adaptive reduction strategy enables the optimization workflow to avoid unnecessary evaluations in regions that are unlikely to contribute to the final Pareto set or optimal-solution discovery. As a result, the effective complexity of the optimization problem is substantially reduced while maintaining access to informative candidate regions.

\subsection{Classification-Based Active Learning}

The refinement stage combines a single XGBoost classifier with the DAGS active-learning query mechanism. The classifier estimates the probability that each candidate belongs to the promising class, whereas DAGS determines which unlabeled candidate should be evaluated next using probability-based uncertainty together with the candidate-space sampling mechanism. The newly obtained oracle label is added to the labeled set, and the XGBoost model is retrained at the next refinement iteration.

For each objective, a benefit-oriented normalized value $z_j(x) \in [0,1]$ is computed so that a larger value is always better:
\begin{equation}
z_j(x) =
\begin{cases}
\dfrac{f_j(x)-f_j^{\min}}{f_j^{\max}-f_j^{\min}}, & \text{for maximization}, \\
\dfrac{f_j^{\max}-f_j(x)}{f_j^{\max}-f_j^{\min}}, & \text{for minimization}.
\end{cases}
\end{equation}
The Pareto-aware proxy components are then defined as
\begin{align}
s_{\mathrm{mean}}(x) &= \frac{1}{m}\sum_{j=1}^{m} z_j(x), \\
s_{\mathrm{worst}}(x) &= \min_j z_j(x), \\
s_{\mathrm{ideal}}(x) &= 1-\frac{\lVert \mathbf{1}-\mathbf{z}(x)\rVert_2}{\sqrt{m}}, \\
s_{\mathrm{geo}}(x) &= \left(\prod_{j=1}^{m}\max\{z_j(x),\varepsilon\}\right)^{1/m}.
\end{align}
The final proxy score is
\begin{equation}
s(x)=0.35s_{\mathrm{mean}}(x)+0.25s_{\mathrm{worst}}(x)+0.25s_{\mathrm{ideal}}(x)+0.15s_{\mathrm{geo}}(x).
\end{equation}
The mean component rewards overall objective quality, the worst-objective component penalizes strongly unbalanced candidates, the ideal-point component favors candidates close to the all-objective ideal vector, and the geometric component promotes balanced simultaneous performance. The score is used only as a dense screening proxy; true Pareto dominance is used exclusively for evaluation and is not used to construct the reduced space.

A percentile threshold is then computed as
\begin{equation}
    \tau = Q_p\left(\{s(x):x\in\mathcal{X}_0\}\right),
\end{equation}
where $Q_p$ denotes the $p$-th percentile of the proxy-score distribution over the complete offline candidate set $\mathcal{X}_0$. In all reported experiments, $p=90$. Binary labels are assigned as
\begin{equation}
    y(x) =
    \begin{cases}
        1, & s(x) \geq \tau, \\
        0, & s(x) < \tau.
    \end{cases}
\end{equation}
This establishes the refinement flow from scalarized scoring to percentile thresholding and binary label assignment.

The initial labeled set contained 16 candidates and was balanced across the positive and negative proxy classes. This controlled initialization prevented the highly imbalanced upper-percentile labeling rule from producing a single-class training set. Both the global percentile threshold and balanced initialization are feasible here because the experiments use offline benchmark datasets with objective values already available for all candidates. In a genuinely online application, proxy scores for unevaluated candidates could not be computed without oracle access, and the labeling strategy would require an online approximation or adaptive threshold.

A single XGBoost classifier $C_{\mathrm{XGB}}$ is used throughout the refinement stage. Given a candidate $x$, the classifier estimates the probability that the candidate belongs to the high-performing class:
\begin{equation}
    p(x)=P(y=1\mid x;C_{\mathrm{XGB}}).
\end{equation}
This probability is used both as a predictive ranking signal and as the basis for uncertainty estimation. Using a single classifier ensures that the refinement mechanism is consistent across datasets and avoids variation caused by combining heterogeneous predictive models.

Prediction uncertainty\cite{ref5} is computed directly from the XGBoost positive-class probability:
\begin{equation}
    u(x) = p(x)\left(1-p(x)\right),
\end{equation}
which is maximized at $p(x)=0.5$, corresponding to candidates located near the current classification boundary.

For oracle querying, DAGS combines model-boundary information with feature-space novelty and local-density information. Candidate novelty is estimated from its distance to the currently labeled samples, while a prediction-space term reflects the distance of its probability prediction from the labels already represented in the labeled set. A nearest-neighbour density term discourages repeated selection of nearly redundant candidates. The selected query is therefore not determined solely by $p(x)(1-p(x))$, but by the density-aware DAGS acquisition rule applied to the current candidate pool.

Pseudo-labels were assigned only to highly confident positive predictions according to
\begin{equation}
    p(x) \geq \lambda_h,
\end{equation}
where $\lambda_h=0.90$. Candidates satisfying this condition were incorporated as pseudo-labeled training samples during subsequent classifier updates.

Pseudo-labeled candidates were used only for subsequent classifier fitting. They were not counted as oracle-evaluated samples and were not transferred to the BO warm-start set unless they had also been explicitly queried. Only oracle-queried candidates were included in $\mathcal{D}_{\mathrm{DAGS}}$.

The reduced candidate set at iteration $k+1$ is defined as
\begin{equation}
\mathcal{X}_{k+1}=\mathcal{X}_{\mathrm{pred}}\cup\mathcal{X}_{\mathrm{model}}\cup\mathcal{X}_{\mathrm{unc}}\cup\mathcal{X}_{\mathrm{conf}}\cup\mathcal{X}_{\mathrm{query}},
\end{equation}

The reduced candidate space is constructed by retaining five complementary categories of candidates: (i) candidates predicted as promising by the classifier, (ii) candidates with high classifier confidence, (iii) candidates with high predictive uncertainty, (iv) high-confidence pseudo-labeled candidates, and (v) all oracle-queried samples. The union of these subsets forms the final reduced search space used by the subsequent warm-start Bayesian optimization stage.

The uncertainty-retention component is percentile-capped: only candidates whose uncertainty lies above the 65th percentile are retained by this criterion. This prevents uncertainty-based preservation from retaining the entire decision-boundary region and thereby undermining the intended space reduction.

This hybrid retention mechanism preserves classifier-positive predictions, candidates with high classifier probability, candidates satisfying the capped uncertainty criterion, high-confidence predictions, and all oracle-queried samples. The combined strategy reduces the probability of eliminating Pareto-relevant candidates during adaptive filtering.

\subsection{Materials Datasets and Objectives}

The proposed framework was evaluated on two distinct optimization benchmarks to assess its scalability, robustness, and generalization capability across different objective landscapes and search-space characteristics.

The first benchmark corresponds to the design optimization of a high-pressure composite vessel and comprises 52,272 Bouligand CFRP laminate configurations generated specifically for this study. The design space is defined by five discrete parameters governing the laminate stacking sequence: starting ply angle ($0{:}5{:}175^\circ$), inter-ply pitch ($5{:}5{:}55^\circ$), ply count ($8{:}1{:}40$), symmetry flag ${0,1}$, and ply-thickness mode ${1,2}$. Full enumeration of the resulting $36 \times 11 \times 33 \times 2 \times 2$ combinations produced 52,272 unique laminate designs.

For each configuration, a CFRP pressure vessel consisting of a cylindrical body with spherical end caps, a diameter of 165~mm, and a half-length of 350~mm was simulated under an internal pressure of 62.5~MPa using Abaqus/CAE. The vessel was modeled using S8R and STRI65 shell elements, with symmetry boundary conditions and unidirectional IM7/8551 orthotropic CFRP plies incorporating Hashin damage initiation~\cite{Hashin}. The resulting material-direction stress components and total vessel-wall thickness were used to define a three-objective minimization problem:
\begin{align}
f_1(x) &= S_{11}(x), \\
f_2(x) &= S_{22}(x), \\
f_3(x) &= T(x),
\end{align}
where $S_{11}$ and $S_{22}$ denote the stress components along the material directions and $T$ represents the total vessel-wall thickness.

The second benchmark consists of a large-scale covalent--organic frameworks (COFs)~\cite{ref16} dataset containing 69,839 candidate structures described by 43 structural, topological, and physicochemical descriptors~\cite{Mercado2018}. These include porosity-related properties, adsorption thermodynamics, geometrical cell parameters, and elemental composition features. The descriptors were generated from atomistic simulations and structural characterization, providing a physically grounded representation of each COF candidate.

For this benchmark, the optimization focuses on methane adsorption performance under pressure-swing adsorption (PSA) operating conditions~\cite{Mercado2018}. The multi-objective optimization problem is defined as
\begin{equation}
\max_x \mathbf{F}(x) = \left[f_1(x), f_2(x)\right],
\end{equation}
where
\begin{align}
f_1(x) &= \mathrm{HighUptake}_{\mathrm{mol}}(x), \\
f_2(x) &= \mathrm{DelCapacity}(x).
\end{align}
Here, $\mathrm{HighUptake}_{\mathrm{mol}}$ denotes the high-pressure methane uptake, while $\mathrm{DelCapacity}$ represents the methane deliverable capacity. Related Bayesian optimization approaches have also been applied to gas-uptake prediction in nanoporous materials~\cite{ref28}. Under the raw maximization formulation, the COF dataset contains 78 true Pareto-optimal solutions, resulting in a more complex multi-objective landscape than the pressure-vessel benchmark.

\subsection{Integrated Optimization Workflow and Evaluation Protocol}

To evaluate the effectiveness of the proposed framework, two optimization configurations were examined: full-space Bayesian optimization and reduced-space Bayesian optimization guided by the proposed adaptive refinement stage. In the baseline full-space setting, Bayesian optimization was directly applied to the complete candidate space without any prior refinement, serving as the primary reference for assessing optimization quality and computational scalability. In the reduced-space setting, the optimization was preceded by the proposed DAGS-based refinement process, which iteratively filters the search space through classification-guided active learning. The resulting reduced candidate subset was subsequently used for Bayesian optimization, with informative samples collected during the DAGS refinement stage transferred as initial observations:
\begin{equation}
    \mathcal{D}^{\mathrm{ref}}_0 = \mathcal{D}_{\mathrm{DAGS}}.
\end{equation}
The optimization process then proceeds iteratively as
\begin{equation}
    \mathcal{D}_t = \mathcal{D}_0 \cup \{(x_i, y_i)\}_{i=1}^{t},
\end{equation}
where $\mathcal{D}_0 = \mathcal{D}^{\mathrm{ref}}_0$ denotes the initialization set inherited from the DAGS refinement stage.

This integration establishes the two-stage refinement-to-optimization workflow illustrated in Figure~\ref{fig:dags-guided-bo-framework}, in which low-cost classifier predictions guide the allocation of expensive objective evaluations.

The refinement stage used 16 initial labeled points and a candidate pool of 8,000 candidates. The positive-class threshold was the 90th percentile, the positive-probability threshold was 0.90, the classifier-probability retention percentile was 50, and the uncertainty retention percentile was 65. The XGBoost classifier was updated for three DAGS iterations on the pressure-vessel benchmark and five DAGS iterations on the COF benchmark.

Both configurations used an identical budget of 150 new BO evaluations, while the DAGS-guided configuration additionally reused the oracle-evaluated samples acquired during the refinement stage as warm-start observations. The resulting warm starts contained approximately 19 queried observations for the pressure-vessel benchmark and 21 for the COF benchmark, compared with five initial observations and 150 BO evaluations for Full BO.

Performance was evaluated using optimization-quality and search-space-retention metrics. For multi-objective tasks, hypervolume (HV)\cite{ref2,ref8,ref30} was used as the primary convergence metric:
\begin{equation}
    \mathrm{HV}_{\mathrm{ret}} = \frac{\mathrm{HV}(\mathcal{X}_k)}{\mathrm{HV}(\mathcal{X}_0)},
\end{equation}
while Pareto preservation and top-performing candidate retention were measured as
\begin{equation}
    P_{\mathrm{ret}} = \frac{|\mathcal{P}_k|}{|\mathcal{P}_0|},
\end{equation}
and
\begin{equation}
    T_{\mathrm{ret}} = \frac{|\mathcal{T}_k|}{|\mathcal{T}_0|}.
\end{equation}


All experiments were implemented in Python using BoTorch, GPyTorch, XGBoost, and scikit-learn under consistent computational settings to ensure fair comparison across all optimization strategies. The complete DAGS-guided refinement and warm-start BO procedure is summarized step by step in Table~\ref{alg:dags-guided-bo}.

\begin{table}[H]
\centering
\caption{Algorithm 1. Overall DAGS-guided Bayesian optimization workflow.}
\label{alg:dags-guided-bo}
\begin{tabularx}{\textwidth}{p{0.07\textwidth}X}
\toprule
Step & Operation \\
\midrule
1 & Input full candidate space $\mathcal{X}_0$ and evaluation budget. \\
2 & Construct a balanced initial labeled set containing high-value and low-value proxy classes, and retrieve their oracle objective values. \\
3 & Compute direction-aware normalized objective values and the Pareto-aware proxy score. \\
4 & Train the XGBoost classifier to distinguish promising and non-promising candidates. \\
5 & Use DAGS to query an uncertainty-guided informative candidate for oracle labeling. \\
6 & Assign pseudo-labels to high-confidence candidates. \\
7 & After the final refinement iteration, construct the reduced candidate space by retaining classifier-positive, high-confidence, uncertain, and queried samples. \\
8 & Repeat refinement until the stopping criterion is reached. \\
9 & Transfer informative samples $\mathcal{D}_{\mathrm{DAGS}}$ to BO. \\
10 & Warm-start qNEHVI Bayesian optimization on $\mathcal{X}_{\mathrm{reduced}}$. \\
11 & Return the final Pareto-front approximation. \\
\bottomrule
\end{tabularx}
\end{table}

\section{Results}

We first evaluate the effect of the adaptive refinement stage on the two benchmark settings by quantifying how much of the original design space is removed and how much useful optimization information is retained. Table~\ref{tab:performance-summary} summarizes the reduction ratio for each dataset together with retention metrics that measure whether high-value regions are preserved, including Pareto-front retention and hypervolume retention. These metrics indicate whether the reduced candidate spaces remain representative of the original optimization landscape before the subsequent Bayesian optimization stage is analyzed in detail.

\begin{table}[H]
\centering
\caption{Summary of DAGS-guided optimization performance across datasets. Values are averaged across ten independent runs. Pareto-AUC improvements are reported relative to full-space Bayesian optimization.}
\label{tab:performance-summary}
\small\begin{tabular}{lcccc}
\toprule
Dataset & Reduction & Pareto Ret. & HV Ret. & Pareto-AUC Imp. \\
\midrule
Pressure vessel & 49.6\% & 69.9\% & 99.6\% & $1562 \rightarrow 2123$ \\
CH$_4$/N$_2$ COFs & 44.2\% & 87.8\% & 99.5\% & $2547 \rightarrow 3199$ \\
\bottomrule
\end{tabular}
\end{table}

All experiments were repeated across ten independent random seeds, and results are reported as mean $\pm$ one standard deviation. No formal claims of statistical significance are made unless supported by an explicitly reported paired test.

\subsection{Adaptive Search-Space Refinement}

The first set of experiments evaluates the ability of the proposed DAGS-guided refinement stage to reduce the search space while preserving the most informative candidate solutions before Bayesian optimization is performed. Rather than directly assessing optimization performance, this analysis investigates whether the adaptive reduction mechanism can eliminate low-value regions without compromising the quality of the remaining candidate space.

\begin{figure}[H]
    \centering
    \includegraphics[width=\textwidth]{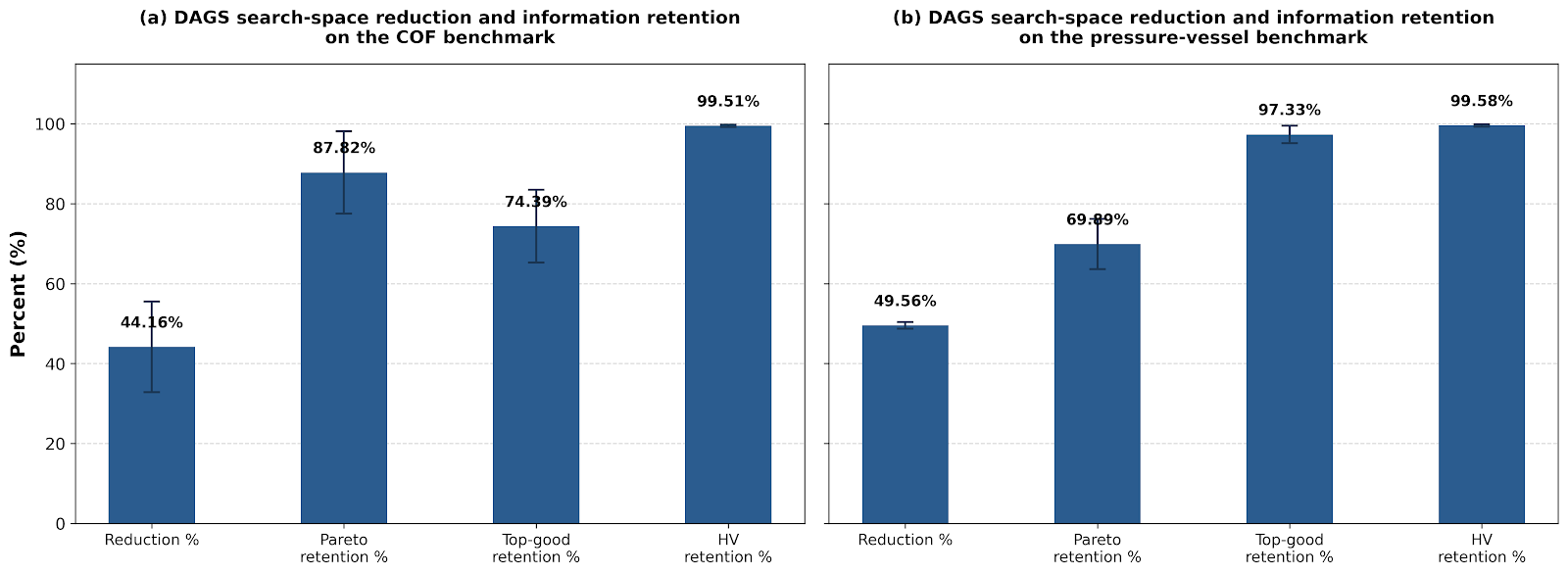}
    \caption{DAGS search-space reduction and information retention for (a) the COF benchmark and (b) the pressure-vessel benchmark. Bars represent the mean over ten independent runs, while error bars indicate one standard deviation.}
    \label{fig:dags-reduction}
\end{figure}

Figure~\ref{fig:dags-reduction} summarizes the average reduction ratio together with the preservation of three complementary quality indicators across ten independent runs for both benchmarks: panel~(a) presents the CH$_4$/N$_2$ COF adsorption dataset, and panel~(b) presents the pressure-vessel benchmark.

For the pressure-vessel benchmark shown in Figure~\ref{fig:dags-reduction}(b), the proposed refinement removed approximately 49.6\% of the original search space while preserving 69.9\% of the true Pareto-optimal solutions, 97.3\% of the top-ranked candidates, and 99.6\% of the original hypervolume. Despite eliminating nearly half of the candidate space, the retained subset maintained almost the entire optimization landscape, indicating that the discarded solutions predominantly corresponded to low-quality design regions.

A similar behavior was observed for the CH$_4$/N$_2$ adsorption optimization problem shown in Figure~\ref{fig:dags-reduction}(a). In this case, the adaptive refinement reduced the search space by approximately 44.2\%, while preserving 87.8\% of the Pareto-optimal solutions, 74.4\% of the top-performing candidates, and 99.5\% of the original hypervolume. Although the Pareto-front geometry differs substantially from that of the pressure-vessel benchmark, the refinement process consistently retained the most informative regions of the objective space.

An important observation is that hypervolume preservation remained above 99\% for both optimization problems, despite the substantial reduction of the candidate space. Since hypervolume simultaneously captures both convergence and diversity of the Pareto front, these results indicate that the proposed refinement removes largely redundant regions while maintaining the overall Pareto-front structure required for subsequent optimization.

Across both datasets, DAGS removed a substantial fraction of the candidates while retaining more than 99\% of the original hypervolume. This indicates that the refinement predominantly removes objective-space redundancy while maintaining an information-rich candidate set for subsequent BO.

\subsection{Hypervolume Convergence}

\begin{figure}[H]
    \centering
    \begin{minipage}[t]{0.49\textwidth}
        \centering
        \includegraphics[width=\linewidth]{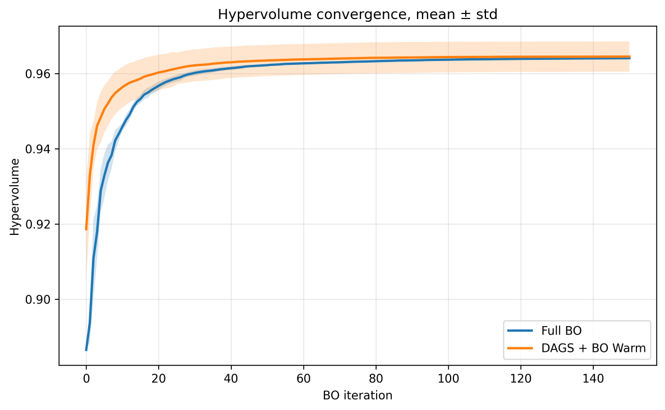}
        \textbf{(a)} Pressure-vessel benchmark
    \end{minipage}\hfill
    \begin{minipage}[t]{0.49\textwidth}
        \centering
        \includegraphics[width=\linewidth]{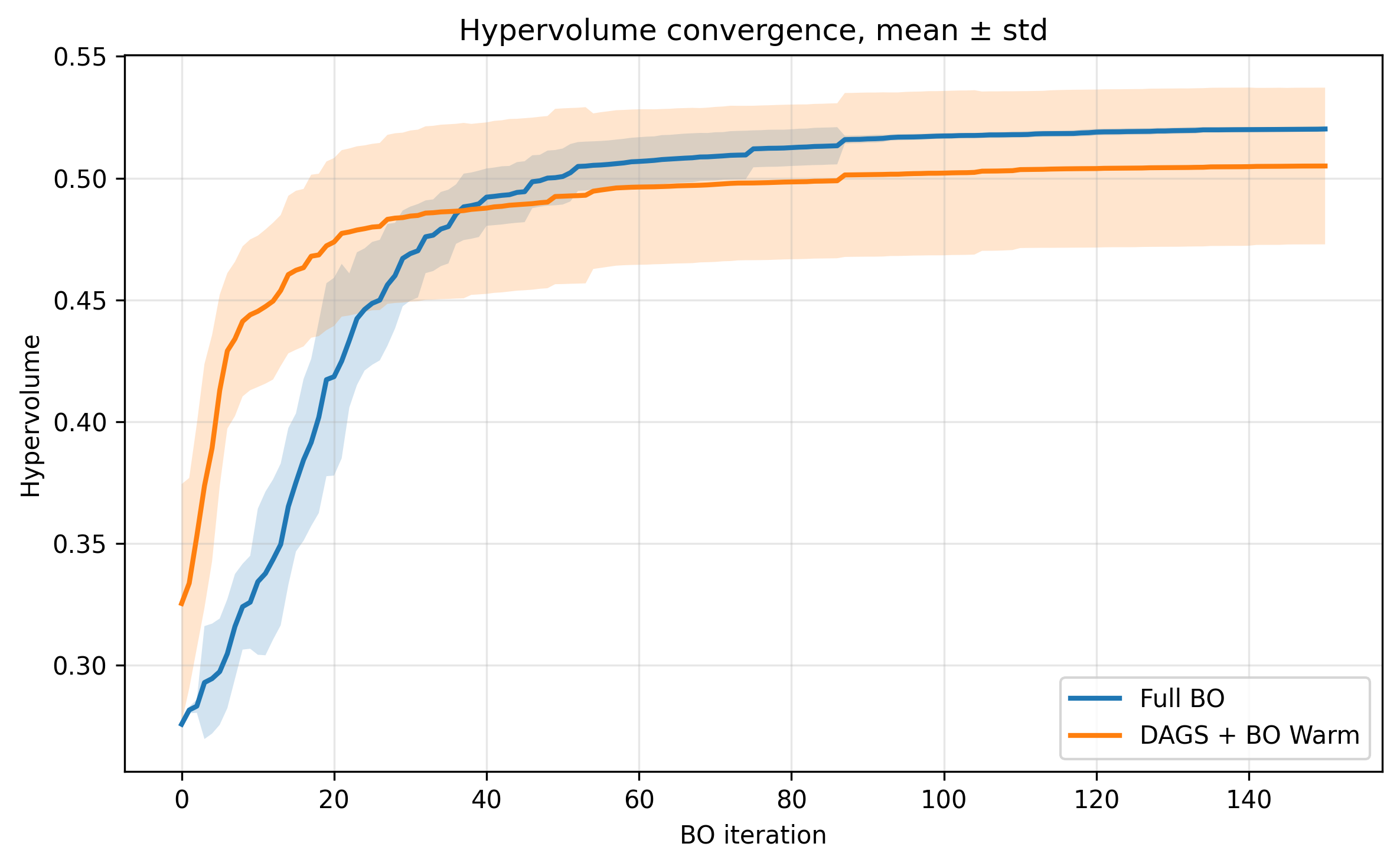}
        \textbf{(b)} COF benchmark
    \end{minipage}
    \caption{Hypervolume convergence of Full BO and DAGS-guided warm-start Bayesian optimization for (a) the pressure-vessel benchmark and (b) the COF benchmark. Curves represent the mean over ten independent runs, while shaded regions indicate one standard deviation. Both methods use 150 new BO evaluations; the warm-start curves additionally reuse oracle-evaluated DAGS samples.}
    \label{fig:hypervolume-convergence}
\end{figure}

The second experiment evaluates the effect of the proposed refinement on the convergence behavior of Bayesian optimization. Figure~\ref{fig:hypervolume-convergence} compares conventional full-space Bayesian optimization and the proposed DAGS-guided warm-start strategy for (a) the pressure-vessel benchmark and (b) the CH$_4$/N$_2$ COF adsorption dataset. Results correspond to the mean performance over ten independent optimization runs, while the shaded regions represent one standard deviation.

For the pressure-vessel optimization problem shown in Figure~\ref{fig:hypervolume-convergence}(a), the proposed framework exhibits faster convergence during the early optimization iterations. Since Bayesian optimization is initialized using informative samples collected during the adaptive refinement stage, the Gaussian process surrogate begins from a substantially more representative approximation of the objective landscape. Consequently, the warm-start optimization reaches comparable hypervolume values after fewer BO-stage iterations. This comparison concerns the 150-evaluation BO stage; the additional oracle evaluations incurred during DAGS refinement are reported separately.

As optimization progresses, both approaches converge towards similar final hypervolume values. Nevertheless, the proposed framework consistently maintains a slight performance advantage while requiring fewer exploratory BO-stage iterations during the early stages of the optimization process. In the pressure vessel benchmark, the DAGS-guided strategy reaches near-plateau hypervolume within roughly the first 20--30 BO iterations, whereas the full-space baseline requires approximately 40--50 iterations to reach comparable values. This behavior indicates that the adaptive refinement primarily accelerates BO-stage convergence rather than merely improving the final solution quality.

The same trend is observed for the CH$_4$/N$_2$ adsorption optimization problem shown in Figure~\ref{fig:hypervolume-convergence}(b). Although the difference in final hypervolume is relatively modest, the proposed warm-start strategy reaches competitive hypervolume values substantially earlier than conventional Bayesian optimization.

These observations demonstrate that transferring informative samples from the adaptive refinement stage enables Bayesian optimization to focus its BO-stage search on promising regions of the design space, allowing faster convergence toward high-quality Pareto approximations within the fixed 150-evaluation BO stage.

\subsection{Pareto-Front Discovery}

While hypervolume measures the overall quality of the obtained Pareto approximation, it does not explicitly quantify the optimizer's ability to identify true Pareto-optimal solutions. Therefore, the next experiment evaluates the cumulative number of true Pareto points discovered throughout the optimization process.

\begin{figure}[t]
    \centering
    \makebox[\textwidth][c]{%
        \begin{minipage}[t]{0.56\textwidth}
            \centering
            \includegraphics[width=\linewidth]{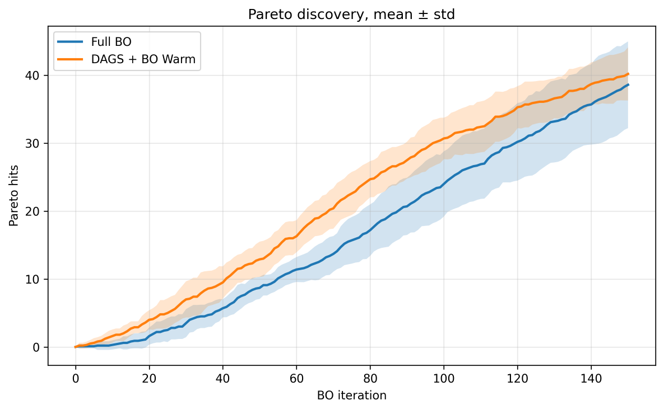}
            \textbf{(a)} Pressure-vessel benchmark
        \end{minipage}%
        \begin{minipage}[t]{0.56\textwidth}
            \centering
            \includegraphics[width=\linewidth]{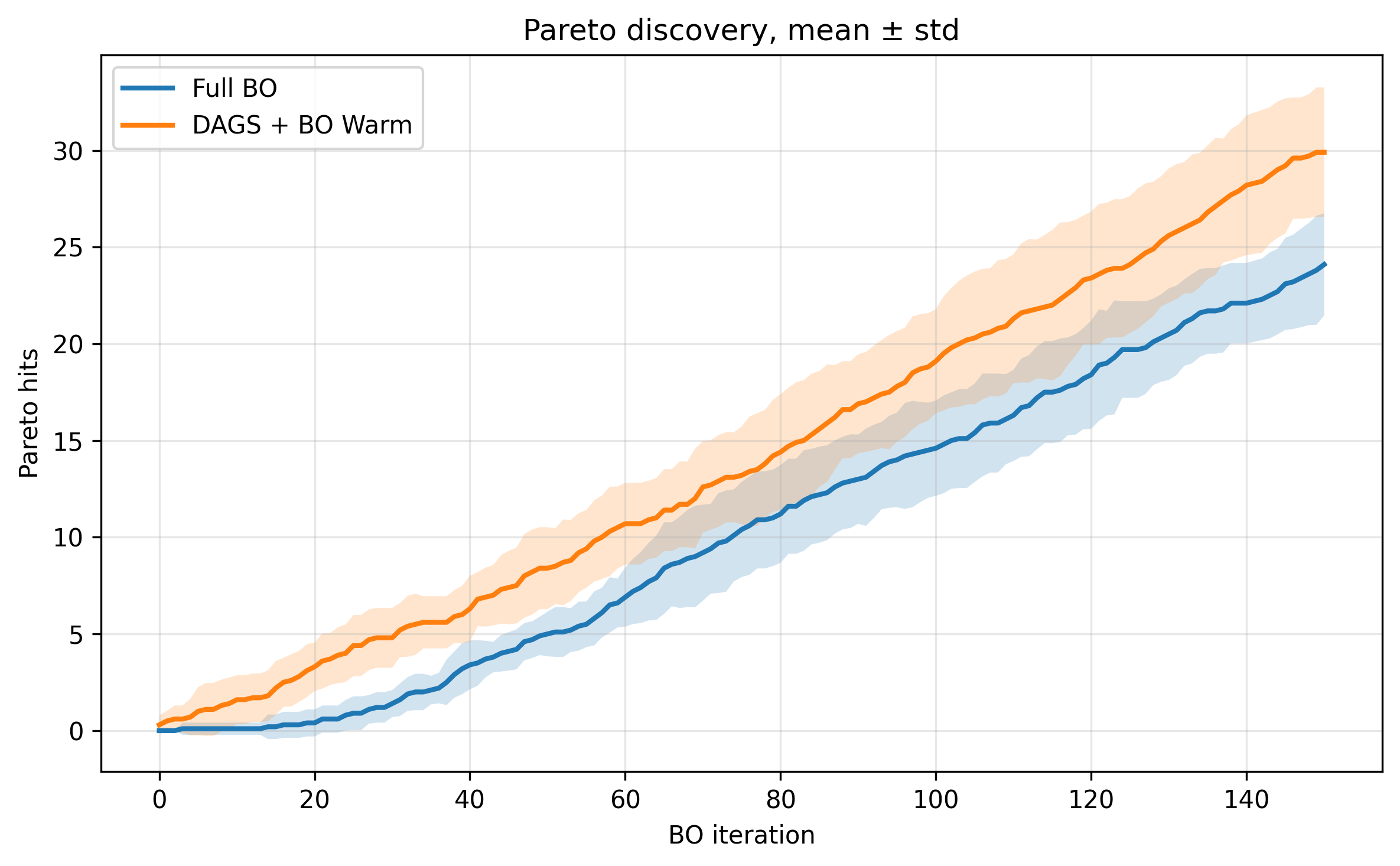}
            \textbf{(b)} COF benchmark
        \end{minipage}%
    }
    \caption{Pareto discovery of Full BO and DAGS-guided warm-start Bayesian optimization for (a) the pressure-vessel benchmark and (b) the COF benchmark. Curves represent the mean cumulative number of discovered Pareto-optimal solutions over ten independent runs, while shaded regions indicate one standard deviation. Both methods use 150 new BO evaluations; the warm-start curves additionally reuse oracle-evaluated DAGS samples.}
    \label{fig:pareto-discovery}
\end{figure}

Figure~\ref{fig:pareto-discovery} presents the average Pareto discovery curves over ten independent runs for (a) the pressure-vessel benchmark and (b) the CH$_4$/N$_2$ COF adsorption dataset. For both optimization problems, the proposed DAGS-guided warm-start strategy consistently discovers Pareto-optimal solutions at a faster rate than conventional Bayesian optimization.

For the pressure-vessel benchmark shown in Figure~\ref{fig:pareto-discovery}(a), the DAGS-guided strategy maintains a higher mean cumulative Pareto-discovery trajectory over most of the evaluation budget. Similarly, in the CH$_4$/N$_2$ adsorption dataset shown in Figure~\ref{fig:pareto-discovery}(b), the proposed method identifies informative Pareto-optimal candidates considerably earlier and continues to outperform the conventional optimizer during most of the optimization trajectory.

This behavior indicates that the adaptive refinement stage successfully concentrates the optimization process around highly informative candidate regions before Bayesian optimization begins. Rather than allocating evaluations to low-value areas of the search space, the optimizer is immediately directed toward regions exhibiting a substantially higher probability of containing Pareto-optimal solutions.

\begin{figure}[t]
    \centering
    \includegraphics[width=\textwidth]{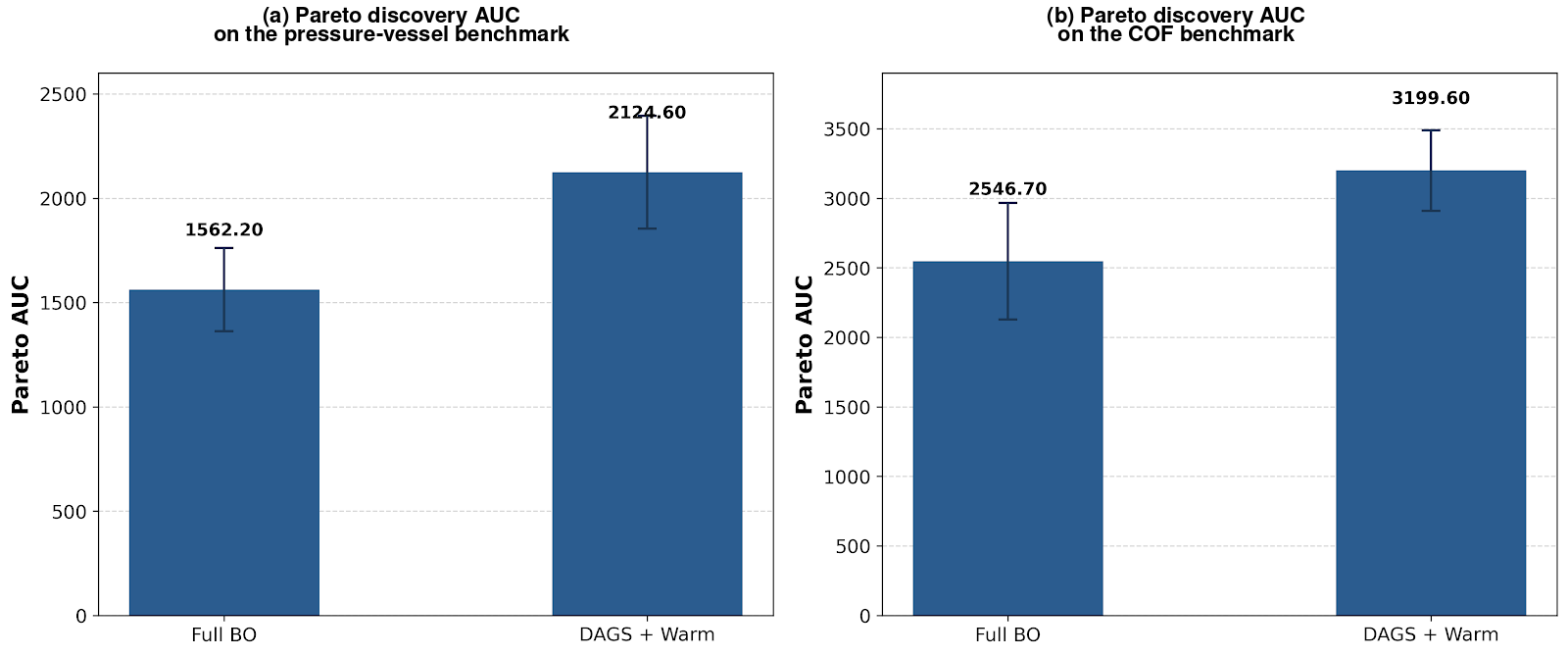}
    \caption{Pareto discovery area under the curve (Pareto-AUC) for (a) the pressure-vessel benchmark and (b) the COF benchmark. Bars represent the mean over ten independent runs, while error bars indicate one standard deviation.}
    \label{fig:pareto-auc}
\end{figure}

To provide a quantitative assessment of the overall optimization trajectory, the area under the Pareto discovery curve (Pareto-AUC) was computed for each optimization run. As illustrated in Figure~\ref{fig:pareto-auc}, the proposed framework consistently achieves larger Pareto-AUC values for (a) the pressure-vessel benchmark and (b) the CH$_4$/N$_2$ COF adsorption problem.

Specifically, the average Pareto-AUC increases from approximately 1562 to 2123 for the pressure vessel benchmark and from 2547 to 3199 for the CH$_4$/N$_2$ adsorption dataset, confirming that the proposed refinement improves Pareto discovery throughout the entire optimization process rather than only at convergence.

Collectively, the hypervolume and Pareto discovery analyses indicate that the proposed refinement improves both optimization speed and solution discovery. Although final hypervolume improvements remain moderate, the substantially larger Pareto-AUC values demonstrate that informative solutions are identified considerably earlier during optimization.

One of the main objectives of this work is to evaluate whether the proposed framework generalizes across optimization problems with substantially different characteristics. The pressure vessel benchmark represents a constrained engineering design optimization problem with three competing objectives, whereas the CH$_4$/N$_2$ adsorption dataset corresponds to a large-scale materials discovery problem involving nearly seventy thousand candidate covalent--organic frameworks and two adsorption-related objectives.

Despite these fundamental differences in search-space structure, objective behavior, and Pareto-front geometry, the proposed framework exhibits remarkably consistent behavior across both domains. In both cases, the adaptive refinement successfully eliminates nearly half of the candidate space while preserving almost the entire hypervolume of the original design space. Furthermore, warm-start Bayesian optimization consistently accelerates convergence and discovers a larger number of Pareto-optimal solutions under an identical budget of 150 new BO evaluations, while additionally reusing oracle-evaluated DAGS samples as warm-start observations.

\section{Discussion}

We presented an active-learning-guided adaptive search-space refinement framework for scalable materials optimization. The approach combines single-classifier XGBoost prediction, DAGS-guided active learning, probability-based uncertainty, high-confidence pseudo-labeling, Pareto-aware proxy scoring, and warm-start multi-objective Bayesian optimization. Across the pressure vessel and CH$_4$/N$_2$ COF benchmarks, the framework reduced the search space by approximately 45--50\% while preserving more than 99\% of the original hypervolume. The DAGS-guided warm-start strategy also improved cumulative Pareto discovery, increasing Pareto-AUC from 1562 to 2123 for the pressure vessel benchmark and from 2547 to 3199 for the COF dataset. These results support adaptive Bayesian search as a promising direction for scalable autonomous materials discovery under limited evaluation budgets.

The proposed framework addresses a key scalability limitation of conventional Bayesian optimization by introducing an adaptive refinement stage that reduces the effective candidate space before expensive optimization. By combining classification-based active learning\cite{ref4,ref5,ref7}, high-confidence pseudo-labeling, and reduced-space BO, the method reallocates computational effort from low-value regions toward regions with higher likelihood of contributing to optimal or Pareto-optimal solutions. The observed improvements originate from the ability of the adaptive refinement stage to bias the Bayesian optimization process toward regions with higher Pareto density while preserving uncertainty around the classification boundary. Consequently, the GP surrogate spends fewer evaluations exploring uninformative regions and allocates more evaluations to candidate solutions with higher expected hypervolume contribution. This explains why the largest gains appear in early convergence and cumulative Pareto discovery rather than only in final hypervolume.

The expected advantages are most pronounced in large candidate pools, where full-space BO may spend many evaluations identifying informative regions. In contrast, the proposed approach uses low-cost XGBoost predictions and pseudo-label-assisted refinement to progressively filter the design space. This strategy can improve BO-stage convergence and search focus while retaining access to high-performing candidates in materials discovery workflows\cite{ref12,ref17,ref22,ref23}. The transfer of informative refinement samples into the BO stage further leverages knowledge collected before expensive optimization, improving search efficiency in large spaces.

The framework also introduces practical trade-offs. Aggressive filtering may remove candidates that later prove valuable, particularly when early labels are noisy or performance thresholds are poorly calibrated. XGBoost probability-based uncertainty, DAGS query selection, and high-confidence pseudo-labeling are used to balance reduction aggressiveness against Pareto-region preservation. Future work should investigate adaptive thresholding, uncertainty-calibrated filtering criteria, and theoretical bounds on Pareto-set retention under classifier-driven space reduction.

The reported convergence advantage should be interpreted as improved BO-stage efficiency rather than a reduction in total oracle evaluations, because the warm-start configuration additionally uses the samples queried during DAGS refinement.

\section{Data Availability}

The datasets used in the current study are available in the zenodo repository, at \url{https://zenodo.org/uploads/21389978}.

\section{Code Availability}

The underlying code for this study is available in the \href{https://github.com/ntagiantas/DAGS-MOBO-Adaptive-Design-Space-Reduction-for-Multi-Objective-Bayesian-Optimization}{GitHub repository}.

\end{document}